%% file: main.tex
\documentclass[conference]{IEEEtran}
\IEEEoverridecommandlockouts

\usepackage{cite}
\usepackage{amsmath,amssymb,amsfonts}
\usepackage{algorithmic}
\usepackage{graphicx}
\usepackage{enumitem}
\usepackage{textcomp}
\usepackage{xcolor}
\usepackage{comment}
\usepackage{caption}
\usepackage{subcaption}
\usepackage{algorithm}
\usepackage{algorithmic}
\usepackage{booktabs}
\usepackage{float} 
\def\BibTeX{{\rm B\kern-.05em{\sc i\kern-.025em b}\kern-.08em
    T\kern-.1667em\lower.7ex\hbox{E}\kern-.125emX}}
\begin{document}

\title{PHANTOM: PHysical ANamorphic Threats Obstructing Connected Vehicle Mobility}
\vspace{-2 in}
\author{\IEEEauthorblockN{Md Nahid Hasan Shuvo and Moinul Hossain}
\IEEEauthorblockA{George Mason University, Fairfax, VA, USA \\
Email:  mshuvo@gmu.edu; mhossa5@gmu.edu }
}
\vspace{-2 in}
\maketitle

\input{body/abstract}

\input{body/introduction}
\input{body/related_works}

\input{body/systemoverview}

\input{body/experiment}
\input{body/conclusion}

\bibliographystyle{IEEEtran}
\bibliography{references}

\end{document}

%% file: body/abstract.tex
\begin{abstract}
Connected autonomous vehicles (CAVs) rely on vision-based deep neural networks (DNNs) and low-latency (Vehicle-to-Everything) V2X communication to navigate safely and efficiently. Despite their advances, these systems remain vulnerable to physical adversarial attacks. In this paper, we introduce PHANTOM (PHysical ANamorphic Threats Obstructing connected vehicle Mobility), a novel framework for crafting and deploying perspective-dependent adversarial examples using \textit{anamorphic art}. PHANTOM exploits geometric distortions that appear natural to humans but are misclassified with high confidence by state-of-the-art object detectors. Unlike conventional attacks, PHANTOM operates in black-box settings without model access and demonstrates strong transferability across four diverse detector architectures (YOLOv5, SSD, Faster R-CNN, and RetinaNet). Comprehensive evaluation in CARLA across varying speeds, weather conditions, and lighting scenarios shows that PHANTOM achieves over 90\% attack success rate under optimal conditions and maintains 60-80\% effectiveness even in degraded environments. The attack activates within 6-10 meters of the target, providing insufficient time for safe maneuvering. Beyond individual vehicle deception, PHANTOM triggers network-wide disruption in CAV systems: SUMO-OMNeT++ co-simulation demonstrates that false emergency messages propagate through V2X links, increasing Peak Age of Information by 68-89\% and degrading safety-critical communication. These findings expose critical vulnerabilities in both perception and communication layers of CAV ecosystems.

\end{abstract}

%% file: body/introduction.tex
\vspace{-0.07 in}
\section{Introduction}
\vspace{-0.01 in}
Autonomous vehicles (AVs) rely on onboard sensors such as cameras and LiDAR for perception. However, due to LiDAR's high cost and challenges of multimodal fusion, many manufacturers like Tesla depend primarily on camera-based perception for object detection and classification. Nonetheless, Camera-based systems are vulnerable to adversarial attacks, a concern amplified further when these vehicles operate as part of connected networks. In particular, Connected Autonomous Vehicles (CAVs) that share perception data through Vehicle-to-Everything (V2X) communication using Basic Safety Messages (BSMs) can propagate information across entire vehicular networks \cite{qayyum2020securing}. Yet, despite this growing interconnectivity, adversarial attacks have primarily been studied in isolated AV settings since Szegedy et al.'s discovery of adversarial examples (AEs) \cite{szegedy2013intriguing}. Existing physical AEs either assume full model knowledge \cite{xue2021naturalae} or create static adversarial elements. Prior studies confirm camera-based AV perception is vulnerable to diverse physical attacks (patches, textures, and patterns) \cite{zhao2025comprehensive}, yet none have explored perspective-dependent adversarial attacks or the network-level impact on the CAV system. This gap leaves systems unprepared for perspective-dependent threats exploiting real-world geometry.

\begin{figure}[t!]
\vspace{0.04in}
\centerline{\includegraphics[width=0.95\columnwidth]{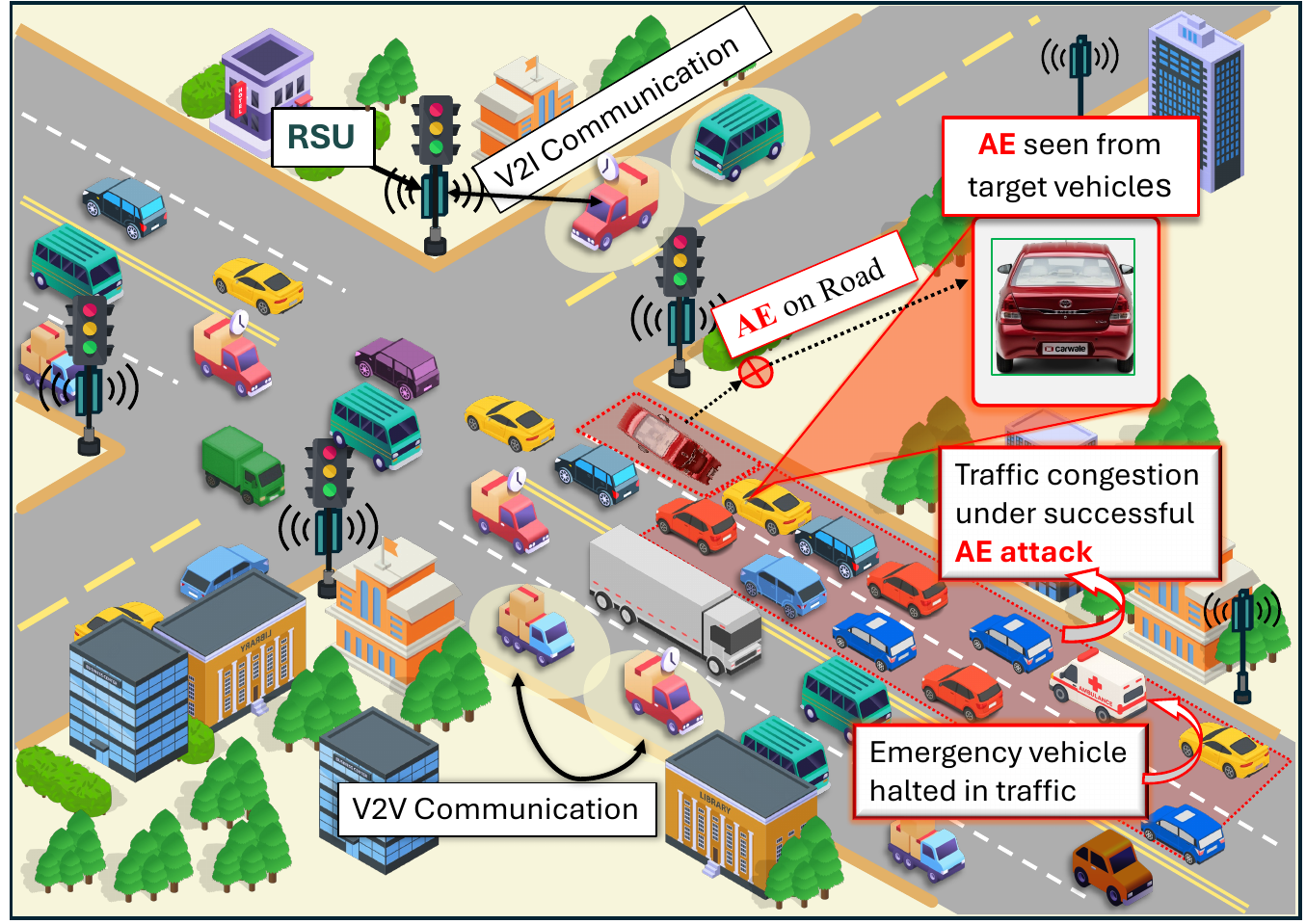}}
\caption{Illustration of PHANTOM physical adversarial attack on CAVs.}\vspace{-0.15in}
\label{fig:vanet_system}
\end{figure}

To address this critical gap, this paper introduces PHANTOM (Physical Anamorphic Threats Obstructing Connected Vehicle Mobility), a novel perspective-dependent physical attack using anamorphic 3D street art to deceive camera-based CAV perception. Unlike prior attacks using conspicuous patches or textures, PHANTOM leverages geometrically distorted patterns that appear as ordinary road markings to human observers, making them highly inconspicuous in real-world settings. The perspective-dependent nature of the attack means it becomes effective only when viewed from specific camera positions corresponding to the ego vehicle’s height and viewing angle, where the distorted artwork transforms into a realistic object. Critically, this paper is the first to demonstrate that a perspective-dependent physical adversarial attack propagates through V2X communication networks via falsified BSMs, triggering cascading network-wide degradation (Fig.~\ref{fig:vanet_system}). Comprehensive evaluation in CARLA \cite{dosovitskiy2017carla} and SUMO-OMNeT++ \cite{behrisch2011sumo,varga2010omnet++} demonstrates the effectiveness and practicality of our proposed attack. 

\vspace{-2 pt}
However, achieving network-wide deception with PHANTOM requires solving four core challenges: \textbf{C1:} Design an anamorphic marking that appears as a realistic object only from a specific viewing distance and angle.  
\textbf{C2:} Determine spatial placement so that the target vehicle reaches the illusion's “sweet spot” with insufficient time for a safe maneuver.  
\textbf{C3:} Ensure environmental robustness so the illusion works across lighting, weather, and vehicle speed variations.  
\textbf{C4:} Operate in black-box settings without access to model internals, producing model-agnostic attacks.
To address these challenges, the key contributions of this paper are as follows:
\vspace{-4 pt}
\begin{itemize}[leftmargin=*]  
    \item We propose a novel attack, PHANTOM, a perspective-dependent physical attack that uses anamorphic art to fool vision systems in autonomous vehicles. The geometry-aware design maintains alignment with the ego vehicle’s viewpoint, producing realistic 3D illusions on flat road surfaces.
    \item We develop an anamorphic AE generation pipeline to create realistic and environmentally resilient attacks. The attack remains robust under varying weather, lighting, and vehicle speeds, and operates in a model-agnostic black-box setting without access to model internals. This attack is validated using CARLA, a physics-based simulator.
    \item We evaluate PHANTOM’s network-level impact through SUMO-OMNeT++ co-simulation, showing that localized perception failures propagate through Vehicle-to-Everything networks and significantly increase the Peak Age of Information (PAoI), which captures the worst-case delay in safety message updates and reveals a critical CAV safety risk beyond the target vehicle by reducing real-time awareness and causing cascading effects across connected vehicles.
\end{itemize}
\vspace{-0.10in}

%% file: body/related_works.tex
\section{Related Works}
\vspace{-1.5 pt}
Following the discovery of neural network vulnerabilities in 2013 by Szegedy et al.\cite{szegedy2013intriguing}, research on adversarial attacks has significantly expanded. In this article, the researcher identified the presence of AEs in the digital domain. Researchers in \cite{goodfellow2014explaining} introduced the fast gradient sign method (FGSM) as a means to create adversarial examples on the MNIST dataset. Authors of the publication \cite{carlini2017towards} proposed three C\&W assaults targeting different distance metrics. This attack was the first published attack to missclassify ImageNet data intentionally.

In \cite{athalye2018synthesizing}, researchers introduced the Expectation Over Transformations (EOT) approach for generating physical AEs, which pioneered the foundation for generating physical AE. In \cite{chen2019shapeshifter}, researchers have shown the physical attack on the TS recognition system, which was based on the Fast-RCNN model. The authors of the \cite{jia2024fooling} showed a framework that can target the digital domain in the physical systems for traffic sign recognition systems. In order to investigate practical security risks more extensively, certain studies \cite{zhao2019seeing}, \cite{chi2024adversarial} have initiated research on physical-world adversarial attacks within the black-box environment, which closely resembles real attack scenarios compared to the white-box setting. Oftentimes, the replica of the white-box model is employed to produce AEs and afterward exploit the transferable nature of these AEs to mislead the targeted black-box models. Recent survey studies on physical adversarial vulnerabilities in autonomous driving systems \cite{zhao2025comprehensive} summarize existing attacks such as patches, textures, and printed perturbations. While authors of \cite{nassi2020phantom} demonstrate that AVs can be fooled by projector-based phantom attacks, their approach is constrained to nighttime operation. In contrast, PHANTOM remains robust across diverse lighting conditions. Moreover, current research largely overlooks geometric transformations, perspective dependency, and the broader network-level impacts of such attacks. PHANTOM addresses these critical gaps by introducing the first perspective-dependent physical adversarial attack based on anamorphic art and evaluating its cascading effects on CAV system.

\vspace{-4 pt}

%% file: body/systemoverview.tex
\section{Proposed Phantom Attack}
\vspace{-0.08in}
Physical adversarial examples (AEs) are real-world modifications that cause a vision model $f(\cdot)$ of an AV to misclassify an input sample $x$ by producing a perturbed input $x'$ that yields $f(x') \not = f(x)$. In this section we describe the system model for PHANTOM. We first present the geometric basis for anamorphic adversarial example generation, then define the attacker capabilities and objectives, and finally describe the full attack pipeline and practical deployment considerations.\vspace{-0.12in}

\begin{figure}[t!]
\centering
\includegraphics[width=0.7\columnwidth]{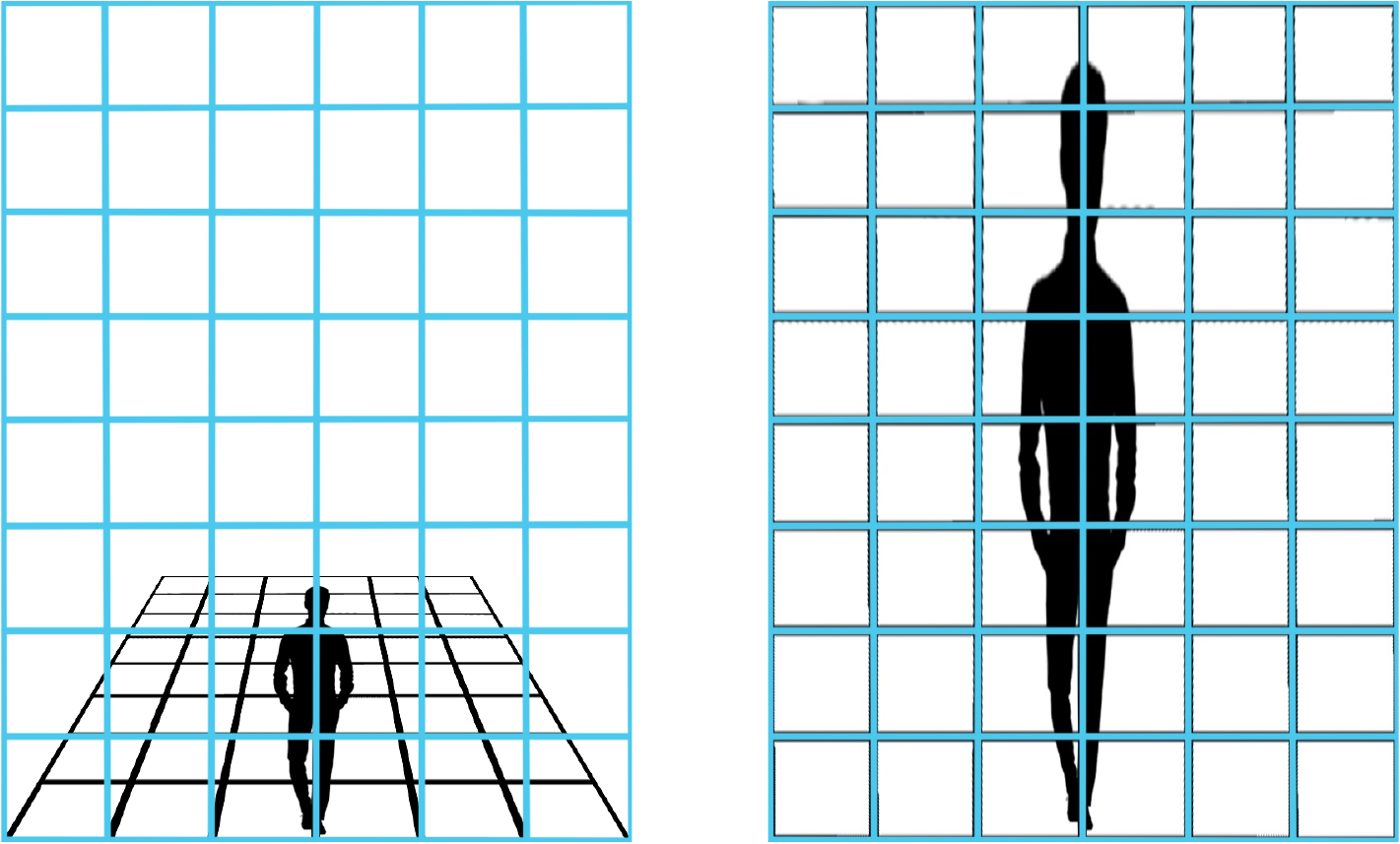}
\caption{Anamorphic art illustration scheme using the grid method.}\vspace{-0.15in}
\label{fig:anamorphic_grid}
\end{figure}
\vspace{-0.03in}
\subsection{Anamorphic Adversarial Example Generation}\vspace{-0.05in}
Anamorphic art produces perspective-dependent illusions that appear distorted on the ground but form realistic 3D shapes from specific viewpoints. PHANTOM leverages this principle to generate adversarial examples that deceive camera-based perception. The generation process subdivides the road plane into a rectangular grid, transforms it into an isosceles trapezoid representing the ego vehicle's viewpoint, and projects the original object image onto this grid, remapping it to the original coordinate plane as of Fig.~\ref{fig:anamorphic_grid}. This creates a road marking that appears realistic when viewed from the target position (sweet spot). The sweet spot is defined by camera height H, viewing distance d, and desired illusion height h. The object width w equals the trapezoid's near edge.


Using geometric projection relations, the far-edge distance $L_g$ and the far-edge width $w_{far}$ of the trapezoid are derived:
\vspace{-0.1in}

\begin{equation}
L_g = \frac{d \cdot h}{H - h}.
\end{equation}\vspace{-0.24in}

The required width of the trapezoid's far edge, $w_{far}$ is obtained from the theorem of similar triangles as:
\vspace{-0.1 in}

\begin{equation}
w_{far} = \frac{w \cdot H}{H - h}.
\end{equation}\vspace{-0.24in}

Now, from $L_g$ and $w_{far}$ we obtain the trapezoid's base angle, $\theta$. 
Which is depended on $w$ and $d$:
\vspace{-0.1in}

\begin{equation}
\theta = \arctan\left(\frac{w}{2d}\right).
\end{equation}\vspace{-0.24in}


This geometric foundation enables the precise construction of anamorphic images to use as physical adversarial examples.
\vspace{-0.25in}

\subsection{PHANTOM Attack Optimization}
Our PHANTOM attack consists of two parts: (1) adversarial example optimization and (2) physical deployment.
\subsubsection{Adversarial Example Optimization}
The optimization stage selects anamorphic parameters $(d,w,h)$ that make the road marking most likely to be misclassified when viewed from the ego vehicle's sweet spot. Given a source image $x_i$ and the geometric transform $T(x_i,d,H,w,h)$, the transformed image is represented as $x' = T(x_i,d,H,w,h)$.

In practice, we employ a two-step search procedure, (1) a coarse grid search over the parameter ranges $D=[d_{\min},d_{\max}]$, $W=[w_{\min},w_{\max}]$, and $B=[h_{\min},h_{\max}]$ identifies promising candidates and (2) a local refinement step using Bayesian optimization  to obtain the best candidates by  maximizing the model loss function $L(\cdot)$:
\vspace{-0.02 in}
\begin{equation}
    \max_{d, w, h} L(f(T(x_i, d, H, w, h)), f(x_i)).
\end{equation}\vspace{-0.1in}

where $f(\cdot)$ denotes the object detector, and each candidate $x'$ is evaluated by a base model (YOLOv2) to estimate transferability to unknown target models. During optimization, practical constraints are enforced: colors and contrasts are limited to printable ranges, and the resulting pattern is regularized to maintain a natural appearance and increase robustness. The complete search and selection process is summarized in Algorithm~\ref{alg:phantom}, and the selected anamorphic image $x'_{\text{best}}$ is forwarded to the Physical Deployment stage.
\vspace{-0.09in}

\begin{algorithm}[th]
\caption{PHANTOM Adversarial Example Generation}
\label{alg:phantom}
\begin{algorithmic}[1]
\begin{small}
\STATE \textbf{Input:} Original image $x_i$, target model $f$, camera height $H$.
\STATE \textbf{Input:} Search ranges for distance $D = [d_{min}, d_{max}]$, apparent width $W = [w_{min}, w_{max}]$, and apparent height $B = [h_{min}, h_{max}]$.
\STATE \textbf{Output:} Optimized anamorphic adversarial image $x'_{best}$.

\STATE Initialize $loss_{best} \leftarrow -\infty$ 
\STATE Initialize $x'_{best} \leftarrow \text{null}$

\FOR{each  $d \in \mathcal{D}$}
    \FOR{each $w \in \mathcal{W}$}
        \FOR{each $h \in \mathcal{B}$}
            \STATE $x'_{current} \leftarrow T(x_i, d, H, w, h)$ 
            \STATE $loss_{current} \leftarrow L(f(x'_{current}), f(x_i))$ 
            
            \IF{$loss_{current} > loss_{best}$}
                \STATE $loss_{best} \leftarrow loss_{current}$
                \STATE $x'_{best} \leftarrow x'_{current}$
            \ENDIF
        \ENDFOR
    \ENDFOR
\ENDFOR

\RETURN $x'_{best}$
\end{small}
\end{algorithmic}
\end{algorithm}
\vspace{-0.10in}

\subsubsection{Physical Deployment}
After obtaining the optimized anamorphic adversarial image $x'_{\text{best}}$, the final step is to physically deploy it on the road surface at the predetermined sweet spot where approaching vehicles will be misled into misclassification. The pattern can be printed as a high-resolution decal or painted using a stencil on asphalt or concrete at suitable deployment sites including intersections, straight roads, or roundabouts where vehicle trajectories are predictable. Deployment requires only basic measurement and printing tools, with geometric parameters (camera height $H$ and approach distance $d$) estimated from public vehicle specifications or field observations. The simplicity and low cost of this process make PHANTOM feasible for real-world execution while maintaining stealth and practicality.\vspace{-0.09in}


\subsection{Threat Model}\vspace{-0.05in}
The attacker operates in a black-box setting with no access to vehicle internals or target model architecture but is aware of common detector types (e.g., YOLOv2). The attacker can physically place anamorphic patterns on road surfaces (intersections, crossings) and estimate geometric parameters (camera height H, approach distance d) from public vehicle specifications. The attackers' objective is to create perspective-dependent illusions that trigger misclassification, emergency braking, and the propagation of misinfromation through the connected network. The attacker cannot interfere with vehicle sensors, software, or communication channels directly. \vspace{-0.02in}




%% file: body/experiment.tex
\vspace{-0.05 in}
\section{Performance Evaluation}\vspace{-0.05 in}
In this section, we are going to discuss our experimental setup and the evaluation of the experiments.\vspace{-0.10in}

\subsection{Experimental Setup}\vspace{-0.05in}
\begin{figure}[!t]
    \centering
    \includegraphics[width=0.9\columnwidth]{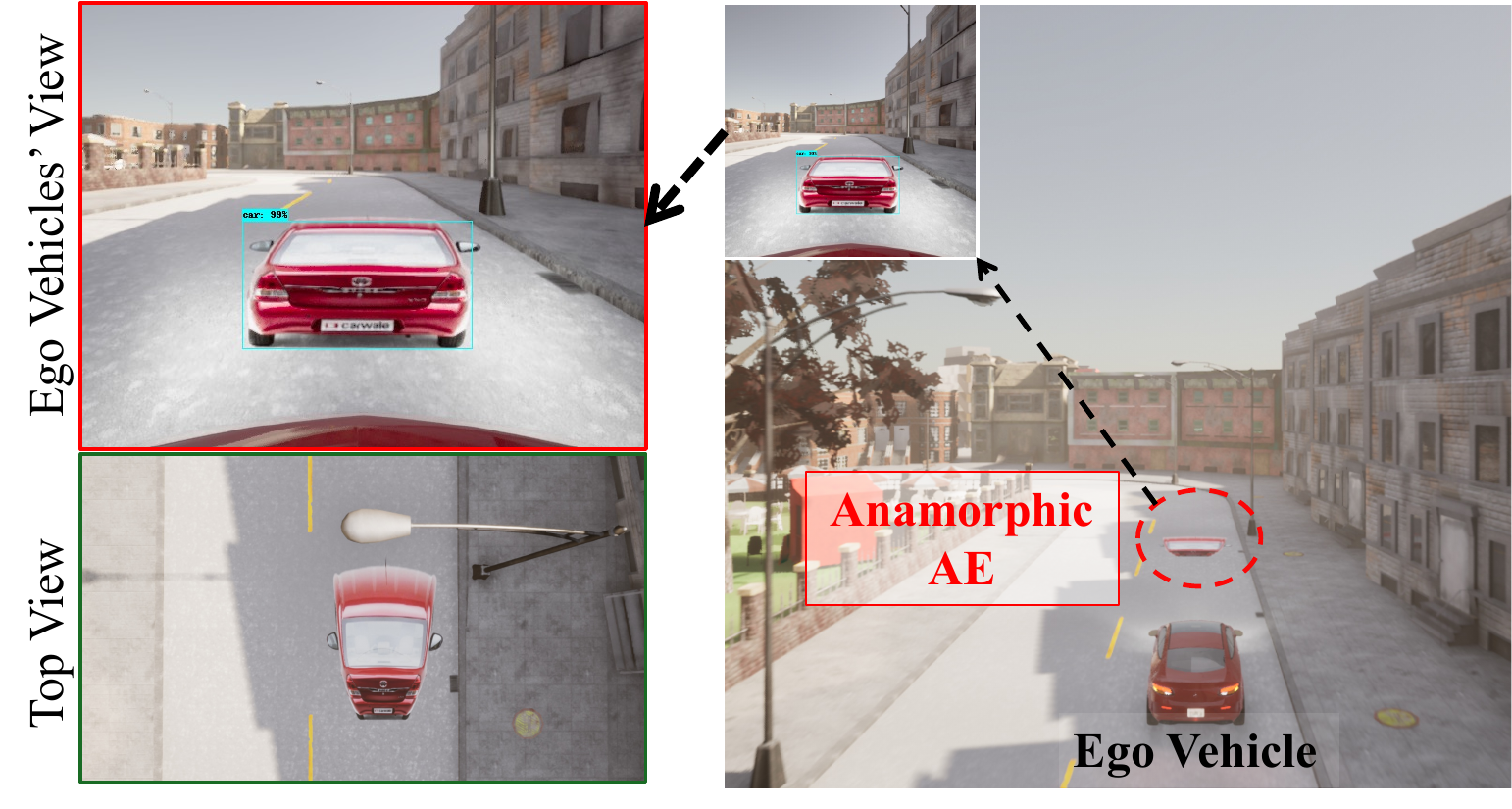}
    \caption{CARLA simulation environment illustrating the PHANTOM attack setup. 
    The top view shows the road with the anamorphic adversarial example (AE) placed as a decal, while the ego-view captures how the illusion appears to the vehicle’s front camera. 
    This configuration enables evaluation of attack activation distance and perception response under realistic driving scenarios.}
    \label{fig:carla_sim}\vspace{-0.18in}
\end{figure}

\subsubsection{\textbf{Data Collection}}
To ensure realistic geometry and texture in the generated anamorphic illusions, the source object images $x_i$ were selected from the CarWale360 dataset \cite{gitdata}, which offers 360° multi-view captures of production vehicles across various viewing angles and lighting environments.

\begin{figure*}[!htb]
    \centering
    \begin{subfigure}{0.48\textwidth}
        \centering
        \includegraphics[width=\linewidth]{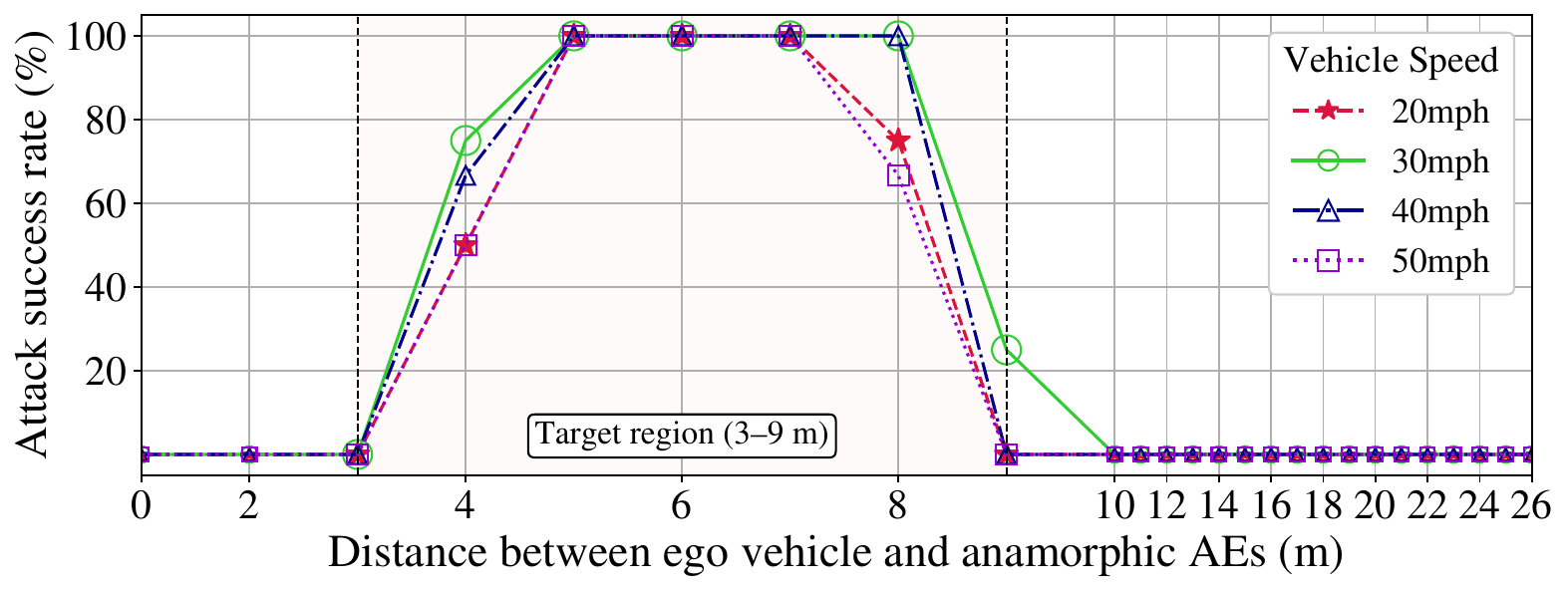}
        \caption{Varying ego vehicle speed.}
        \label{fig:asr_speed}
    \end{subfigure}
    \hfill
    \begin{subfigure}{0.48\textwidth}
        \centering
        \includegraphics[width=\linewidth]{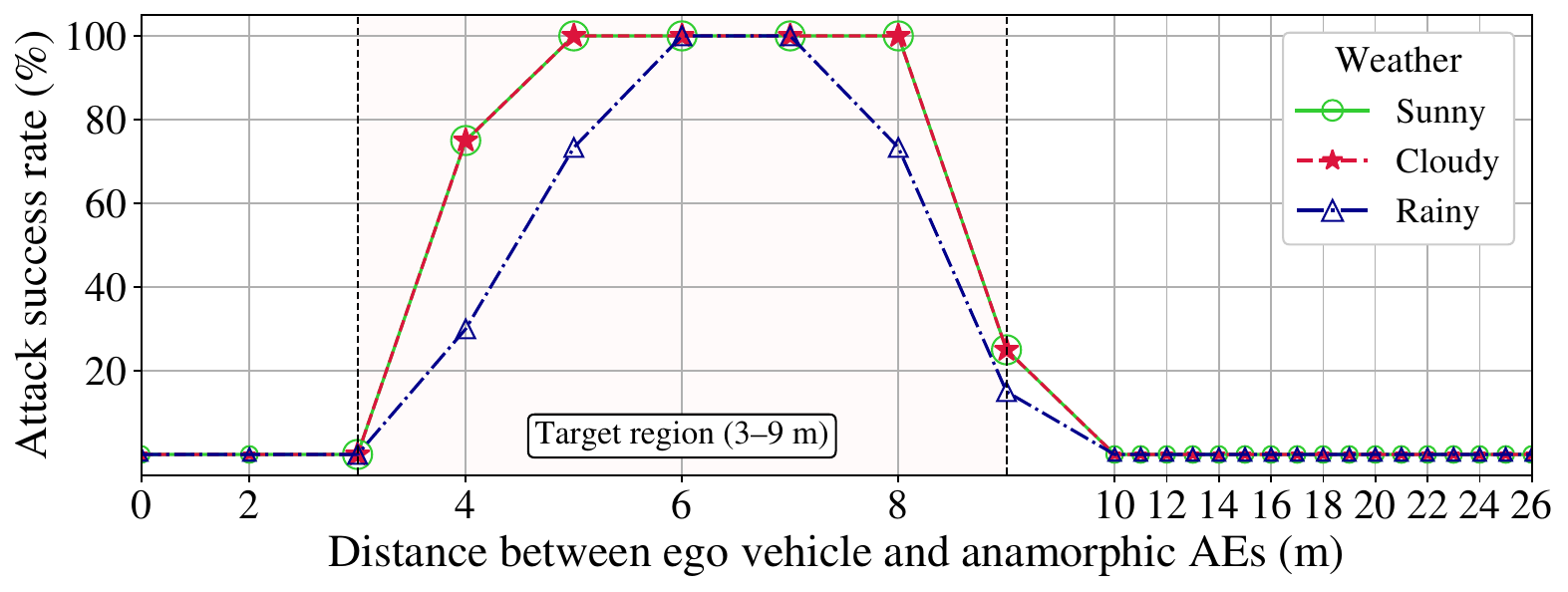}
        \caption{Different weather conditions.}
        \label{fig:asr_weather}
    \end{subfigure}
    \begin{subfigure}{0.48\textwidth}
        \centering
        \includegraphics[width=\linewidth]{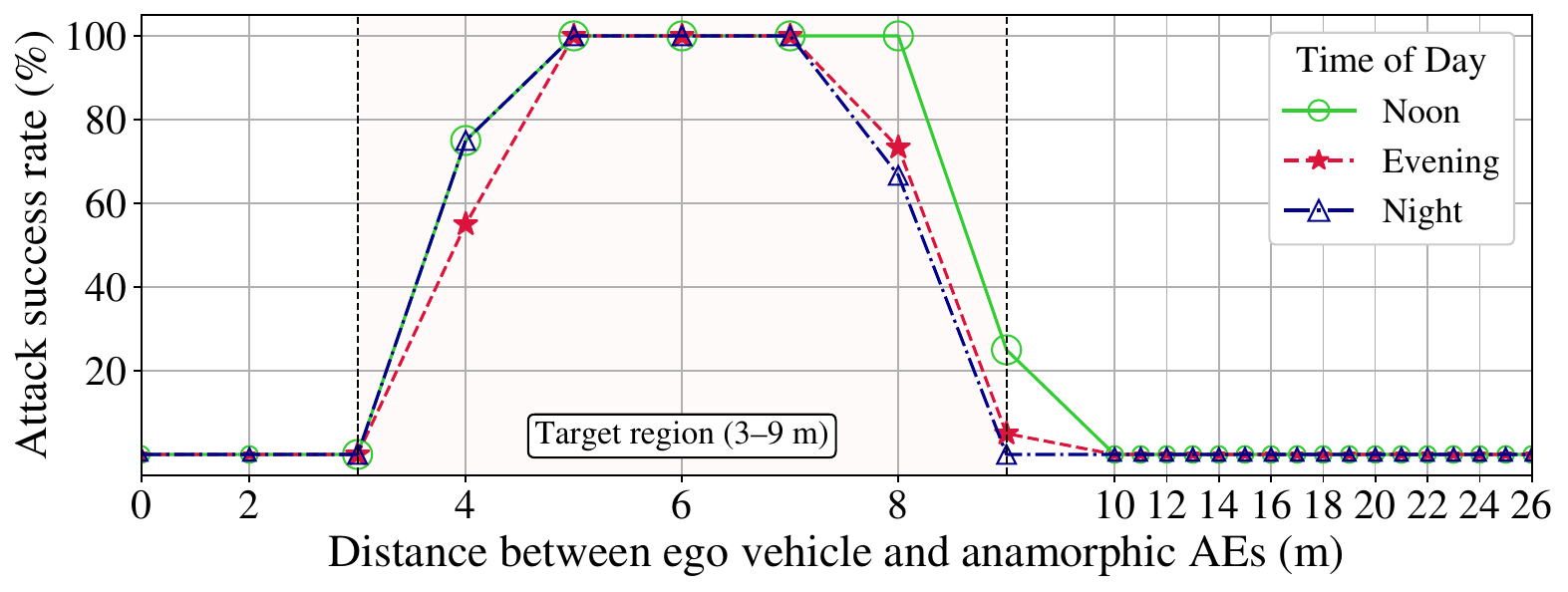}
        \caption{Different times of the day.}
        \label{fig:asr_light}
    \end{subfigure}
    \hfill
    \begin{subfigure}{0.48\textwidth}
        \centering
        \includegraphics[width=\linewidth]{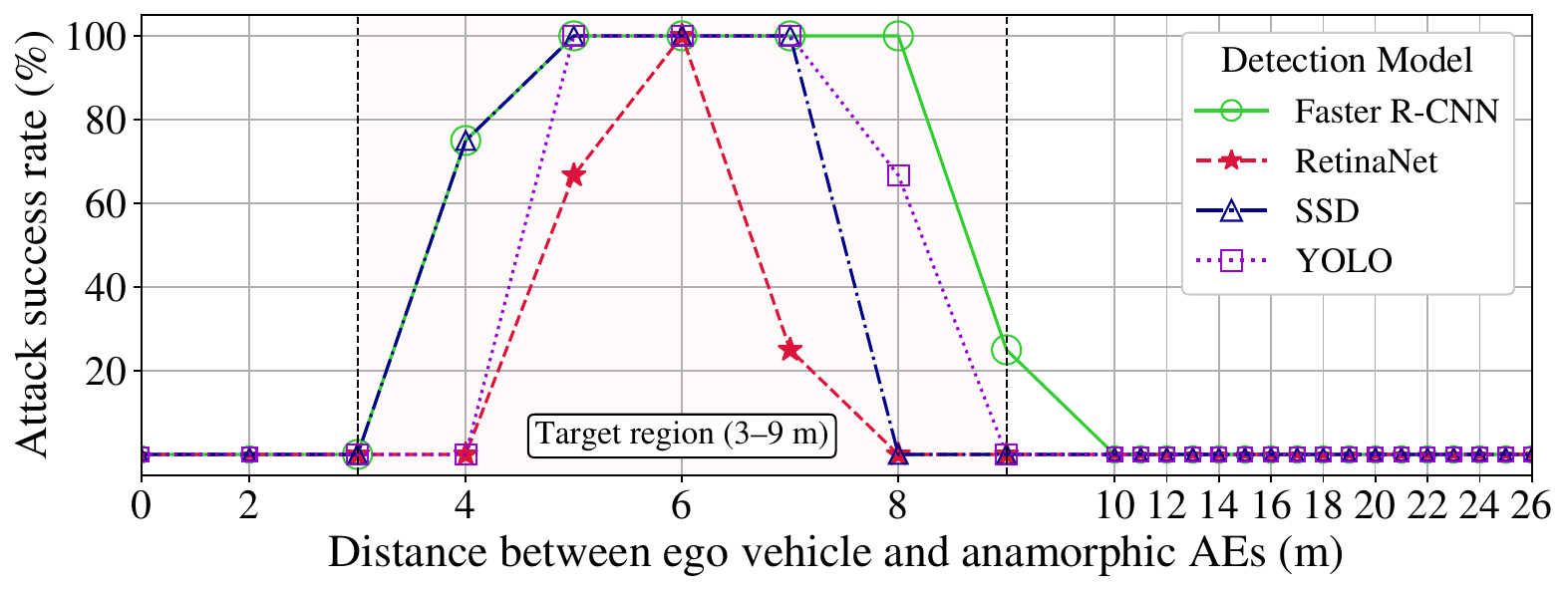}
        \caption{Different detection models.}
        \label{fig:asr_models}
    \end{subfigure}\vspace{-0.05in}
    \caption{Attack success rate vs. distance under different conditions: (a) varying ego vehicle speed, (b) weather, (c) lighting, and (d) detection model.}\vspace{-0.25in}
    \label{fig:asr_2x2}
\end{figure*}


\subsubsection{\textbf{Object Detectors for Evaluation}}
We use multiple object detectors to evaluate PHANTOM across diverse detection architectures. During optimization, we employ YOLOv2 as the surrogate model because of its lightweight structure, fast inference speed, and accessibility of intermediate features, which make it suitable for repeated black-box evaluations during parameter search. For evaluating, we use four widely adopted detectors: YOLOv5, SSD, Faster R-CNN, and RetinaNet as the main detector on the ego vehicle in CARLA. These models represent both one-stage (YOLOv5, SSD, RetinaNet) and two-stage (Faster R-CNN) detection paradigms, covering a broad spectrum of modern perception designs. Using this diverse set enables us to evaluate PHANTOM’s transferability and robustness across model families that differ in architecture, feature extraction depth, and region-proposal mechanisms.\vspace{-0.02in}

\subsubsection{\textbf{Real-World Attack Simulation}}
We evaluate PHANTOM using two complementary simulation environments. 

First, the perception-level experiments are conducted in the CARLA physics-based simulator (Fig.~\ref{fig:carla_sim}) to replicate realistic camera perspectives, vehicle motion, and environmental conditions. To evaluate environmental robustness, we test under multiple lighting and weather conditions (clear, cloudy, and rainy) and across different vehicle speeds (20 to 50~mph). CARLA provides accurate sensor modeling, allowing us to evaluate PHANTOM's attack success rate under realistic driving conditions. Fig.~\ref{fig:carla_sim} shows the PHANTOM attack on CARLA, where target misdecte the AE with 99\% accuracy.

Second, to simulate network behavior for CAVs under our attack scenario, we employed SUMO (Simulation of Urban MObility) and OMNeT++ with IEEE 802.11p for V2X communication. While SUMO controls vehicle mobility, OMNeT++ handles data exchange between Roadside Units (RSUs) and vehicles, mirroring real-world CAV networks. We simulate both benign and attacked scenarios with varying vehicle densities to observe how perception failures cascade through the network and trigger cascading failures via BSMs.\vspace{-0.15in}
\subsection{PHANTOM Attack Evaluation}\vspace{-0.1 in}

All models are implemented in PyTorch and evaluated on NVIDIA GPUs. CARLA simulations run with vehicle camera parameters matching typical production cameras (field of view: 90°). SUMO traffic scenarios are seeded with realistic route distributions and communication patterns based on urban traffic flow data. For distance-based analysis, we sample 5 positions uniformly within each 1 m distance bin and repeat each configuration 5 times with different random seeds, yielding 25 measurements per bin. Our evaluation uses two key metrics: Attack Success Rate (ASR) and Area Under the Curve (AUC). ASR denotes the percentage of frames within each bin in which the adversarial illusion successfully causes object misclassification while maintaining detection accuracy above 80\%, which ensures that the attack is evaluated only under confident model predictions. AUC quantifies the overall attack effectiveness during the critical attack range (3-9 m) as the ratio of cumulative ASR values to the maximum possible ASR values, reflecting the consistency and strength of the attack across distances up to the critical range where the illusion begins to take effect. 

\subsubsection{\textbf{Attack Success in CARLA}}
We evaluate PHANTOM in CARLA across varying vehicle speeds (20-50 mph) to measure attack success under realistic urban driving conditions. Fig.~\ref{fig:asr_speed} reveals a critical distance-speed relationship. AEs begin deceiving detectors when ego vehicles cross the 10 m threshold and achieve 100\% ASR before reaching 7 m across all speeds. This phenomenon results from speed-dependent mechanisms: at low speeds (20 mph), cameras capture sufficient frames for the illusion to develop in the detector's temporal buffer; at high speeds (50 mph), motion blur and rapid perspective changes cause detectors to lock onto the formed illusion without temporal context for verification. For moderate speeds (30-40 mph), the attack initiates from 8 m, providing only 0.4-0.6 seconds of reaction time, which is insufficient for safe maneuvering (minimum 1.5 seconds required) per NHTSA standards \cite{NHTSA_CORE_2018}. At 50 mph, this window shrinks to 0.18 seconds, forcing emergency braking that poses rear-end collision risks. This validates that anamorphic projection exploits the critical gap between detection and safe maneuvering.\vspace{-0.05in}

\textbf{Insight.} \textit{PHANTOM creates an unavoidable safety dilemma by positioning the attack at the "point of no return" for autonomous vehicles. Regardless of vehicle speed, the temporal window between first detection and collision risk is too narrow for safe evasive maneuvers, forcing hard braking that can trigger rear-end collisions in multi-vehicle scenarios.}

\subsubsection{\textbf{Robustness Across Environmental Conditions}}
To evaluate the environmental resilience of PHANTOM, we tested the attack under diverse weather (sunny, cloudy, rainy) and lighting (noon, evening, night) conditions in the CARLA simulator. These conditions represent common real-world scenarios that can alter brightness, contrast, and surface reflection, all of which affect vision-based perception. Fig.~\ref{fig:asr_2x2}(b)--(c) illustrate the variation of attack success rate (ASR) with respect to ego-vehicle distance across these environments. Under sunny and cloudy conditions, PHANTOM maintains nearly perfect performance, achieving 100\% ASR within the 7--8\,m activation range. Rainy conditions slightly degrade performance due to water reflections and motion blur, reducing AUC to 0.56 but still achieving a successful attack at 7\,m. Lighting variations also have minimal impact—evening and night scenarios maintain 50--100\% ASR within 4--7\,m—demonstrating that PHANTOM’s geometry-based illusions remain robust under diverse environmental perturbations. The quantitative summary in Table~\ref{tab:scene} further confirms these findings.\vspace{0.02in}

\begin{table}[!t]\vspace{0.08in}
\centering
\caption{Attack performance under different environmental scenarios.}\vspace{-2 pt}
\label{tab:scene}
\resizebox{0.85\columnwidth}{!}{%
\begin{tabular}{lccccc}
\hline
\textbf{Scenario} & Sunny & Cloudy & Rainy & Evening & Night \\ \hline
AUC & 0.714 & 0.714 & 0.560 & 0.619 & 0.631 \\
Peak Attack Start (m) & 8 & 8 & 7 & 7 & 7 \\ \hline
\end{tabular}%
}
\end{table}\vspace{-2 pt}
\vspace{-2 pt}
\begin{table}[!t]
\centering
\caption{Attack performance across different detection models.}\vspace{-2 pt}
\label{tab:models}
\resizebox{0.85\columnwidth}{!}{%
\begin{tabular}{lcccc}
\hline
\textbf{Models} & Faster R-CNN & YOLO & SSD & RetinaNet \\ \hline
Highest Accuracy & 0.99 & 0.99 & 0.95 & 0.88 \\
AUC & 0.714 & 0.524 & 0.536 & 0.274 \\
Peak Attack Start (m) & 8 & 7 & 7 & 6 \\
\hline
\end{tabular}
}
\end{table}

\textbf{Insight.} \textit{PHANTOM maintains high effectiveness across different environmental scenarios, demonstrating that geometric perspective manipulation is more robust than texture-based adversarial patches, which degrade significantly under lighting and weather variations.} 

\subsubsection{\textbf{Transferability and Black-box Performance}}
To assess model-agnostic robustness, PHANTOM is optimized exclusively on YOLOv2 and tested against four unseen detectors (YOLOv5, SSD, Faster R-CNN, RetinaNet), demonstrating strong black-box transferability in Fig.~\ref{fig:asr_models}. All four exhibit consistent vulnerability despite never being exposed during optimization. Faster R-CNN achieves the highest ASR profile (100\% from 9m), while YOLOv5 and SSD reach 100\% ASR from 7 m. RetinaNet achieves 100\% ASR only at the 6 m with below 70\% elsewhere. However, RetinaNet's base detection accuracy is significantly lower (65-75\%) compared to other models (85-95\%). PHANTOM exploits shared mid-level feature representations (edges, textures, spatial arrangements) universal to CNN-based detectors, confirming robustness under realistic black-box conditions where attackers lack access to target model internals. The detailed numerical results are presented in Table~\ref{tab:models}. Faster R-CNN exhibits the highest AUC (0.71) and earliest activation distance (8 m), followed closely by YOLO and SSD. RetinaNet shows the lowest attack sensitivity due to its weaker baseline accuracy, not due to lack of attack robustness. These findings confirm PHANTOM’s strong transferability across heterogeneous model architectures.

\textbf{Insight.} \textit{The high transferability across diverse detectors reveals that PHANTOM can impose serious attacks on different vision architectures due to the fact that all current vision-based systems share the same feature extraction procedure.}

\begin{figure}[!t]
\centering
\includegraphics[width=0.9\columnwidth]{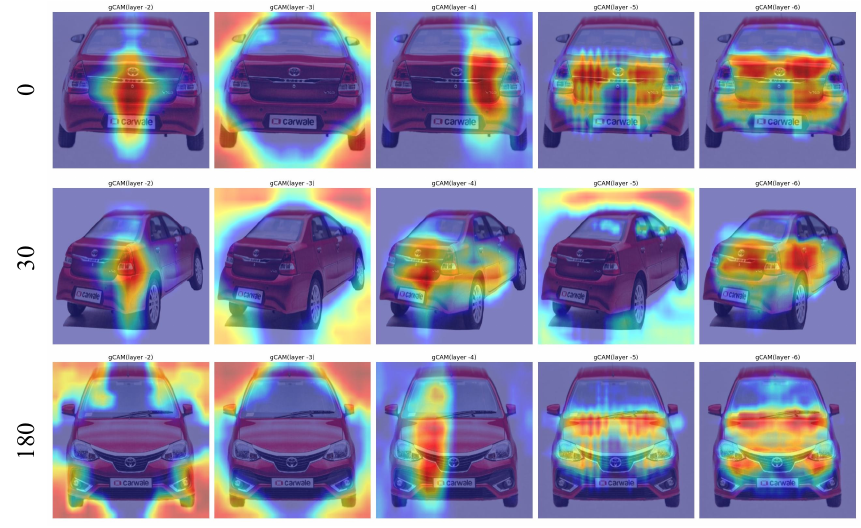}
\caption{Grad-CAM visualization of YOLOv5 under PHANTOM attack.}\vspace{-0.15in}
\label{fig:attn_mapall}
\end{figure}
\begin{figure}[!t]
\centering
\begin{subfigure}[t]{0.45\columnwidth}
    \centering
    \includegraphics[width=\columnwidth]{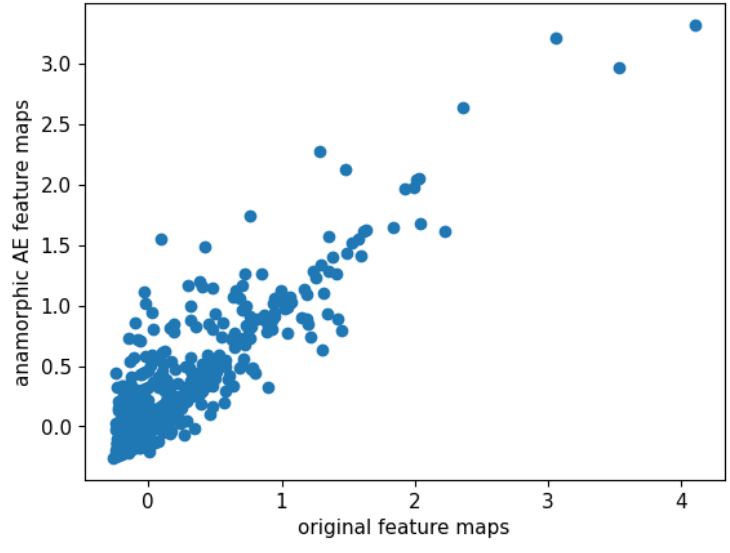}
    \caption{Correlation of feature maps.}\vspace{-0.1in}
    \label{fig:attn_map_a}
\end{subfigure}
\hfill
\begin{subfigure}[t]{0.48\columnwidth}
    \centering
    \includegraphics[width=\columnwidth]{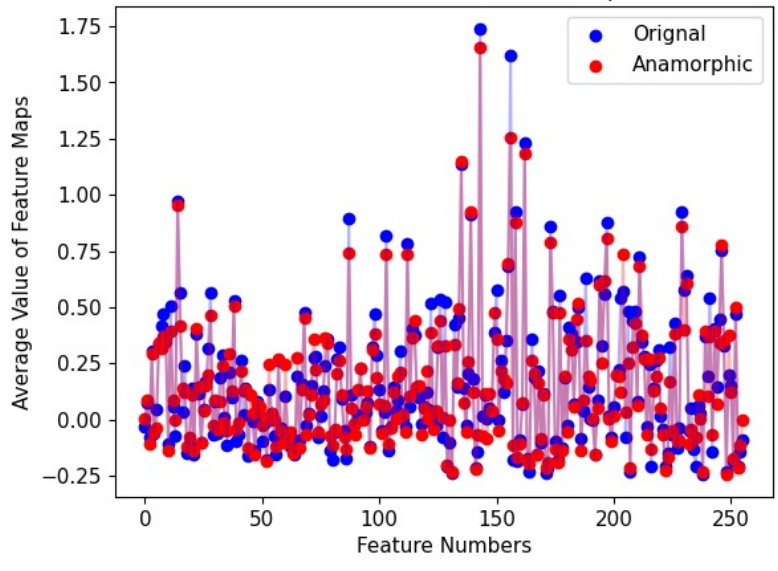}
    \caption{Difference in feature maps.}\vspace{-0.1in}
    \label{fig:fmap_rel_b}
\end{subfigure}
\caption{Relation between the feature maps of benign and anamorphic image.}\vspace{-0.15in}
\label{fig:combined}
\end{figure}

\subsubsection{\textbf{Attention Mapping and Feature Correlation}}
To understand why PHANTOM deceives the detector, we perform Grad-CAM visualization~\cite{Selvaraju_2017_ICCV} on YOLOv5. As shown in Fig.~\ref{fig:attn_mapall}, the attention maps concentrate on the wheel and rear-edge textures of the projected illusion, indicating that detectors rely on locally coherent visual cues for object recognition. To further analyze this behavior, we examine the feature maps of both benign and adversarial images across different network depths. The early layers capture low-level edges and shapes, while deeper layers encode high-level semantic structures. We apply Global Average Pooling to each layer and conduct a feature-map correlation analysis. As shown in Fig.~\ref{fig:attn_map_a}, the correlation coefficient between benign and PHANTOM images reaches approximately 0.97, and Fig.~\ref{fig:fmap_rel_b} further demonstrates that the feature values of the original and anamorphic images are nearly identical. This high similarity explains why YOLOv5 interprets anamorphic illusions as real vehicles—the attack preserves mid-level convolutional patterns that the model associates with genuine objects, thereby bypassing normal feature-based discrimination. These observations clarify how PHANTOM achieves high ASR even without altering any layers of the deep learning models.


\textbf{Insight.} \textit{This analysis reveals that PHANTOM replicates mid-level CNN features to deceive the autonomous vehicle.}

\begin{figure}[!t]
\centering
\begin{subfigure}[t]{0.49\columnwidth}
    \centering
    \includegraphics[width=\columnwidth]{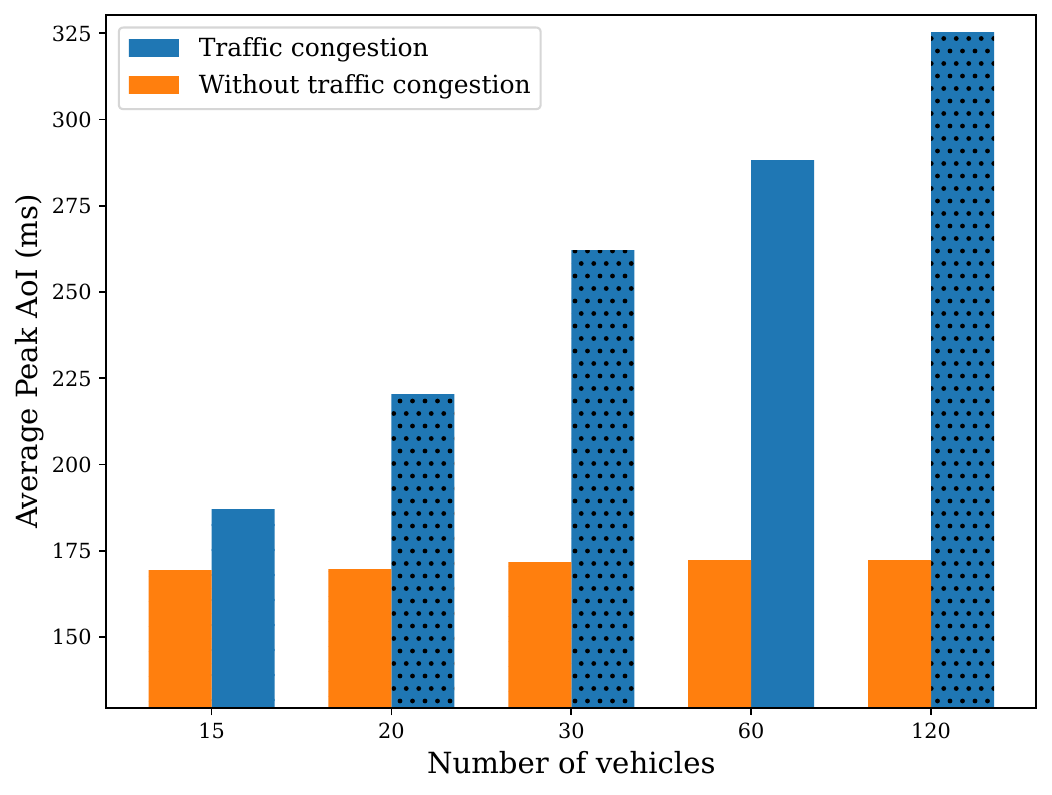}
    \caption{Vehicle density vs. PAoI.}
    \label{fig:net1_l}
\end{subfigure}
\hfill
\begin{subfigure}[t]{0.49\columnwidth}
    \centering
    \includegraphics[width=\columnwidth]{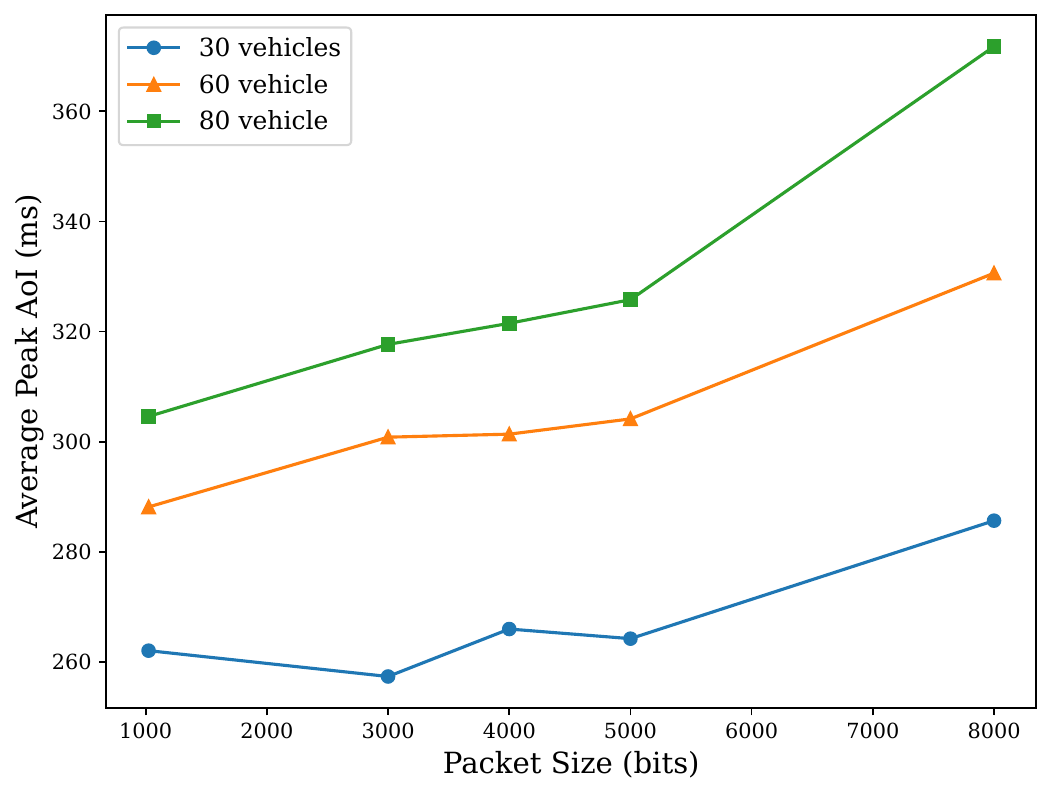}
    \caption{Packet Size vs. PAoI.}
    \label{fig:net1_r}
\end{subfigure}
\caption{Impact of adversarial attacks on connected vehicles.}
\vspace{-0.15in}
\label{fig:net1}
\end{figure}

\subsubsection{\textbf{System-level Impact in SUMO–OMNeT++}}
We extend the analysis to connected-vehicle networks using SUMO–OMNeT++ co-simulation. When a vehicle misclassifies a PHANTOM illusion as an obstacle, it triggers the SAE J2735 emergency protocol, generating high-priority BSMs that propagate to all vehicles within the 300 m RSU communication range via DSRC (IEEE 802.11p)~\cite{sae_j2735}. This initiates a three-stage cascade: (1) hard braking (2) emergency BSM flooding, and (3) reactive braking and route recalculation by surrounding vehicles, creating traffic congestion. Fig.~\ref{fig:net1_l} illustrates the impact on PAoI. It shows PAoI increases dramatically during PHANTOM-induced congestion: 68\% higher with 60 vehicles (287 ms from 170 ms) and 89\% higher with 120 vehicles (321 ms from 170 ms). This stems from emergency BSMs transmitted at 10 Hz (vs. 1 Hz normal), overwhelming RSU capacity. Moreover, Fig.~\ref{fig:net1_r} shows PAoI increases exponentially as packet size grows from 1000 to 8000 bits. At 60 vehicles, PAoI rises from 260 to 340 ms, a 31\% increase. This network congestion confirms that single perception errors cascade into network-wide degradation affecting data freshness and cooperative driving efficiency, which can hinder the propagation of safety-critical information  from real accidents, emergency vehicles.

\textbf{Insight.} \textit{PHANTOM represents a multi-layer attack that a single strategically placed pattern can reduce an entire traffic network's situational awareness to unsafe levels.}

\begin{figure}[t!]
\centering
\includegraphics[width=0.9\columnwidth]{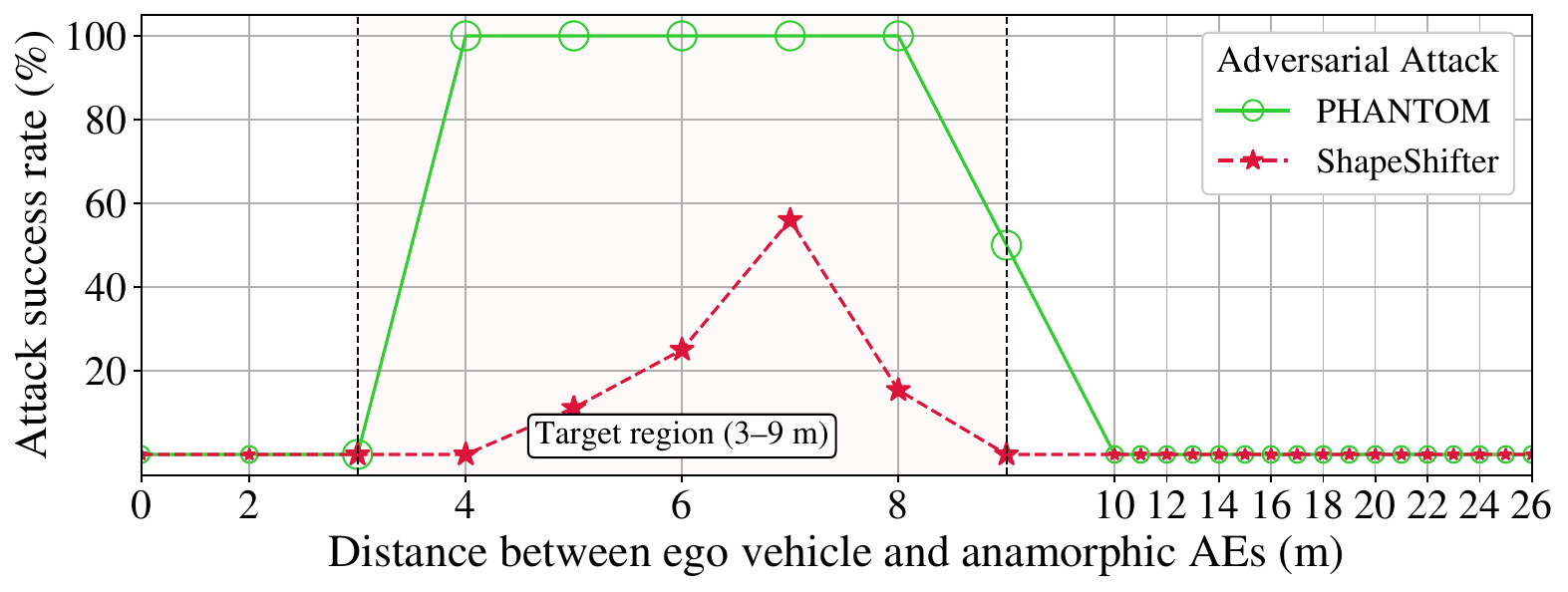}
\caption{Attack success rate vs. distance with varying vehicle speeds, PHANTOM achieves 0.71 AUC, whereas ShapeShifter achieves 0.15 AUC.}
\vspace{-0.15in}
\label{fig:compare}
\end{figure}

\subsubsection{\textbf{Comparison with Baseline}}
We compare PHANTOM to ShapeShifter \cite{chen2019shapeshifter}. While ShapeShifter perturbs existing object surfaces (stop signs), producing targeted misclassification or disappearance, PHANTOM produces false-positive vehicle detections via anamorphic road art. Fig.~\ref{fig:compare} shows that in CARLA environment, ShapeShifter achieved 0.15 AUC. In contrast, PHANTOM achieved 0.71 AUC, demonstrating a 4.7× more robust and sustained attack within the target band. This result shows that PHANTOM's method of generating phantom objects is significantly more effective than ShapeShifter's method of modifying existing ones.

\subsubsection{\textbf{Defense against PHANTOM Attack}}
While PHANTOM exposes vulnerabilities in camera-based CAV perception, existing defenses remain constrained by practicality. Multi-modal fusion with LiDAR (905 nm ToF) or radar (77 GHz FMCW) can validate 3D geometry but incurs hardware cost, calibration complexity, and computational overhead. Temporal filtering via Kalman tracking detects unrealistic motion but fails during abrupt maneuvers. Network-level consensus among vehicles and RSUs mitigates false-positive propagation at the expense of latency. Adversarial training cannot span the full geometric parameter space of our PHANTOM attacks. Collaborative perception offers resilience by cross-view validation, yet occlusion and limited coverage persist \cite{badjie2024adversarial}.
\vspace{-0.1in}

%% file: body/conclusion.tex
\section{Conclusion}\vspace{-0.02in}
This paper presented \textbf{PHANTOM}, a novel physical adversarial attack employing anamorphic art to deceive vision-based perception in connected autonomous vehicles. PHANTOM achieves over 90\% attack success rate and increases PAoI by up to 89\% in connected networks. The attack remains robust across weather, lighting, and detector variations, demonstrating strong black-box transferability. These results reveal a critical vulnerability linking perception and communication layers in CAV ecosystems. Future work will focus on developing geometry-aware, cooperative defenses integrating redundant sensing, BEV-map fusion, and cross-vehicle validation to ensure resilient and real-time perception integrity. \vspace{-0.05in}